\documentclass[a4paper, 10pt, conference]{IEEEtran}   

\usepackage{FG2023}

\FGfinalcopy 

\IEEEoverridecommandlockouts                              

\usepackage{graphicx}
\usepackage{amsmath}
\usepackage{booktabs}
\usepackage{multirow}
\usepackage{amssymb}
\usepackage{pifont}
\newcommand{\xmark}{\ding{55}}%
\usepackage{comment}
\usepackage[hyphens]{url}  

\def\FGPaperID{****} 

\title{\LARGE \bf Are Face Detection Models Biased?}

\author{\parbox{16cm}{\centering
    {\large Surbhi Mittal$^1$, Kartik Thakral$^1$, Puspita Majumdar$^{1,2}$, Mayank Vatsa$^1$ and Richa Singh$^1$}\\
    {\normalsize
    $^1$IIT Jodhpur, India, $^2$IIIT Delhi, India}}
}


\begin{document}

\IEEEoverridecommandlockouts
\IEEEpubid{\makebox[\columnwidth]{979-8-3503-4544-5/23/\$31.00 ©2023 IEEE\hfill} \hspace{\columnsep}\makebox[\columnwidth]{ }}

\ifFGfinal
\thispagestyle{empty}
\pagestyle{empty}
\else
\author{Anonymous FG2023 submission\\ Paper ID 0078 \\}
\pagestyle{plain}
\fi

\maketitle

\begin{abstract}
The presence of bias in deep models leads to unfair outcomes for certain demographic subgroups. Research in bias focuses primarily on facial recognition and attribute prediction with scarce emphasis on face detection. Existing studies consider face detection as binary classification into `face' and `non-face' classes. In this work, we investigate possible bias in the domain of \textit{face detection through facial region localization} which is currently unexplored. Since facial region localization is an essential task for all face recognition pipelines, it is imperative to analyze the presence of such bias in popular deep models. Most existing face detection datasets lack suitable annotation for such analysis. Therefore, we web-curate the Fair Face Localization with Attributes (F2LA) dataset and manually annotate more than 10 attributes per face, including facial localization information. Utilizing the extensive annotations from F2LA, an experimental setup is designed to study the performance of four pre-trained face detectors. We observe (i) a high disparity in detection accuracies across gender and skin-tone, and (ii) interplay of confounding factors beyond demography. The F2LA data and associated annotations can be accessed at \url{http://iab-rubric.org/index.php/F2LA}.



\end{abstract}


\section{Introduction}
\label{sec:intro}

\noindent With deep learning models/algorithms becoming the norm of the modern AI systems, it has become essential to evaluate these algorithms (and systems) thoroughly to minimize any adverse impact on the society. The incorporation of \textit{bias} in the algorithms is one primary issue that has been highlighted in the literature~\cite{drozdowski2020demographic, singh2022anatomizing}. Research has shown that the performance of deep learning algorithms vary for people with different attributes such as gender and skin-tone subgroups under a variety of settings \cite{amazon, propublica}. For instance, it has recently been observed that the automatic face-cropping algorithm of Twitter \textit{favors young and lighter-skinned people} over others~\cite{Twitter}. With increasing instances of these issues, it is of paramount importance to include fairness as one of the metrics for evaluation of these algorithms.

\begin{figure}       
  \centering {\includegraphics[scale = 0.40]{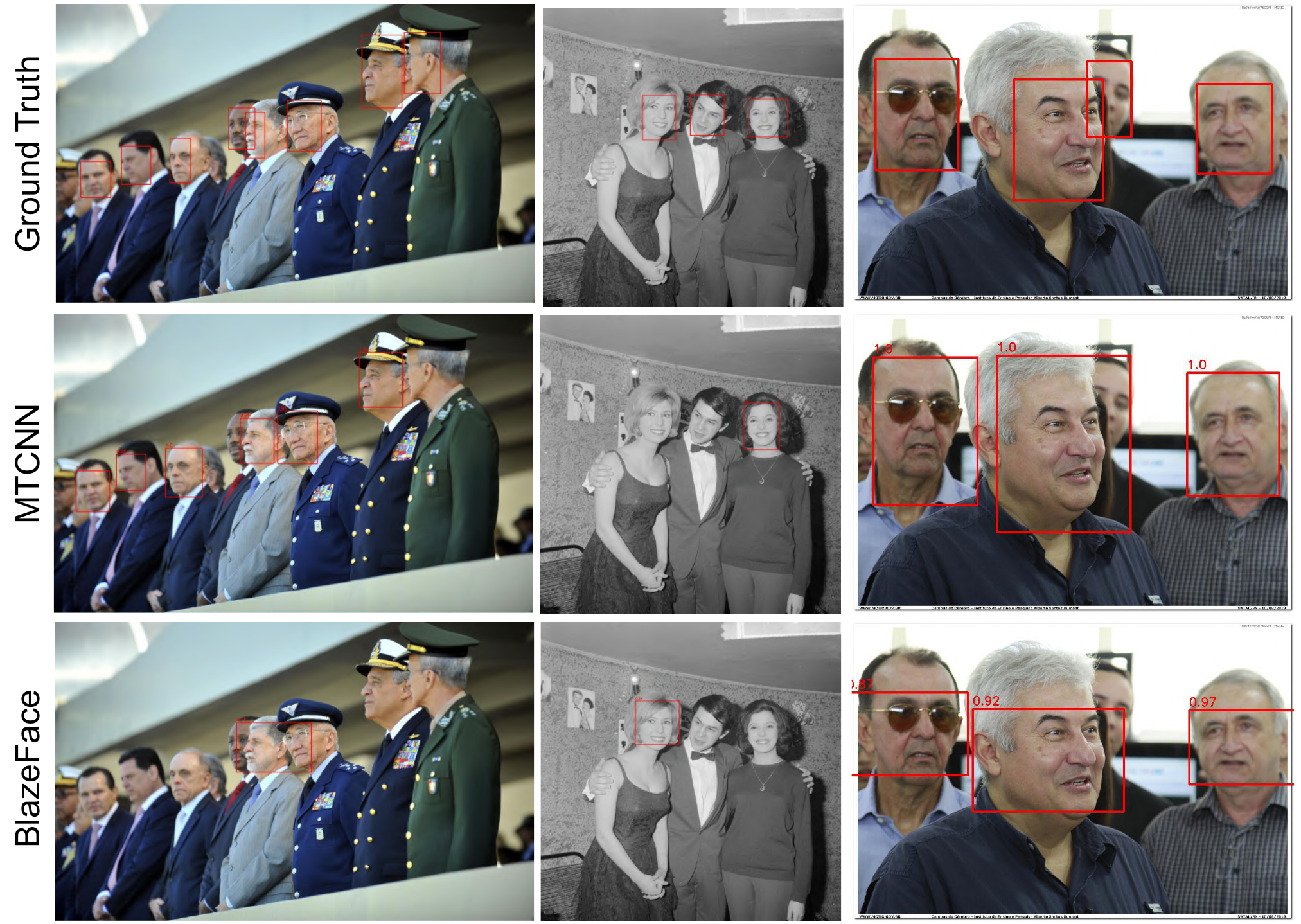}}  
    \caption{Face detection models failing to detect faces.}
    \label{fig:bias_samples}
\vspace{-8pt}
\end{figure}

Fairness of facial analysis algorithms is being studied in the literature for the last few years \cite{majumdar2021unravelling, robinson2020face, aaai2020, singh2022anatomizing}. Most of the research efforts have been concentrated towards establishing awareness towards bias in face \textit{recognition} systems and a number of datasets have been proposed for the same \cite{karkkainen2021fairface, wang2019racial}. However, limited research work has studied the impact of face \textit{detection} which forms an important part of the recognition pipeline and failure in which can lead to incorrect decisions (Fig. \ref{fig:bias_samples}). To the best of our knowledge, none of the existing studies on bias in face detection focus on bounding box localization. As the first contribution of this work, we analyze different facial detectors to understand if they exhibit any biased behavior.


\noindent The presence of bias in deep models has been attributed to non-demographic factors (such as variation in pose, illumination, and image quality), as well as more complex, demographic factors such as race, gender, and skin tone~\cite{kortylewski2019analyzing}. To study biased behavior, datasets with extensive annotations corresponding to different attributes are required which are lacking in existing databases. To facilitate the study of bias for this research, the second contribution is that we have collected images from the web and curated the \textit{Fair Face Localization with Attributes (F2LA)} dataset, with annotation of ten attributes per face along with bounding box localization. Using the F2LA database, we perform a detailed analysis on four face detectors and observe that different confounding factors play an important role in decision making. 
%

\begin{figure*}[]      
  \centering {\includegraphics[width=\textwidth]{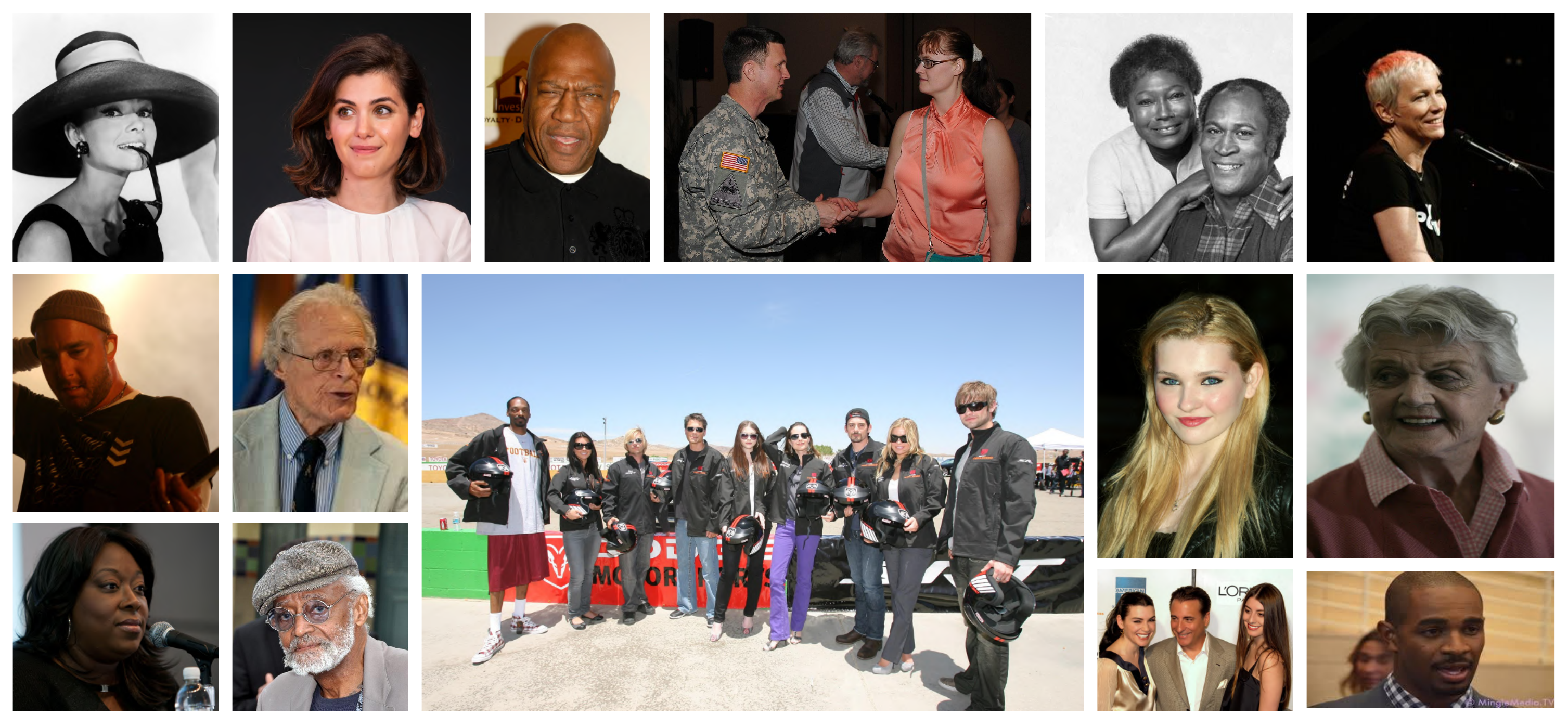}}
    \caption{Sample images of the proposed F2LA dataset with diversity in gender, skin-tone, age, illumination, occlusion and background.}
    \label{fig:db_samples}
\vspace{-8pt}
\end{figure*}

\section{Related Work}

\noindent Facial analysis models have shown superlative performance over the last few years \cite{blazeface,retinaface,arcface,lcsee, mtcnn}. However, biased predictions have been reported in various facial analysis tasks such as face detection, attribute prediction, and face recognition. The seminal paper by Buolamwini et al.~\cite{gendershades} showed the biased behavior of commercial gender classifiers against darker-skinned females. Over the years, multiple studies have been performed in facial recognition and attribute prediction~\cite{majumdar2021attention, singh2022anatomizing}. However, demographic bias in face detection has received significantly less attention. Datasets containing extensive annotations for demographic subgroups as well as non-demographic attributes are required for studying bias. However, popular benchmark databases for face detection including AFW~\cite{afw}, PASCAL face~\cite{pascalface}, FDDB~\cite{fddb}, and WIDER FACE~\cite{widerface} datasets do not contain annotations pertaining to both demographic and non-demographic factors. An existing database for evaluation of bias in faces includes the PPB (Public Parliaments Benchmark) dataset~\cite{gendershades}\footnote{The PPB dataset was requested but was under legal review.}.  While PPB contains face images of individuals from three African countries and three European countries, the dataset contains samples in constrained settings with only one face per image.

\noindent Further, limited approaches to mitigate such bias have been proposed. Amini et al.~\cite{amini2019uncovering} proposed a novel algorithm for mitigating hidden bias in face detection algorithms. The proposed algorithm uses a variational autoencoder to learn the latent structure and the learned latent distributions re-weight the importance of certain data points during training. However, the face detection problem in the paper is formulated as a binary classification problem where each image contained one face. Modern face detection models provide bounding box annotations corresponding to the faces in a given image along with a confidence score for each detected face. The proposed F2LA dataset is an evaluation set containing 1200 images with bounding box localization as well annotations for 10 attributes corresponding to each face for fairness analysis.


\section{Experimental Design}
\noindent In order to quantify bias in face detection, we require a dataset with annotations for bounding box localization as well as for attributes such as demographic subgroups. The proposed \textbf{Fair Face Localization with Attributes (F2LA) dataset} is employed for studying the behavior of face detectors. This section describes the F2LA dataset and details of the experimental protocol designed for analysis.


\subsection{F2LA Dataset}
\noindent We web-curate the \textit{F2LA} dataset containing images of people in unconstrained settings. \textit{Facial localization information for 1774 faces} in 1200 images is manually annotated along with annotations for \textit{10 attributes per face}. These attributes include information pertaining to gender, skin tone, apparent age, facial orientation, blur, illumination, color properties, facial hair, occlusion and background information. These attributes assist in evaluating bias present in existing models across various factors, and provide insights into the fairness of face detection algorithms. The F2LA data and associated annotations can be accessed at \url{http://iab-rubric.org/index.php/F2LA}.

\noindent \textbf{Data Collection and Properties:} The images are collected by crawling the internet for face images with CC-BY licenses. Variation across multiple attributes is ensured during collection. Samples of the dataset are shown in Fig. \ref{fig:db_samples}. Besides localizing each face with a bounding box, several other attributes have been annotated. The details of the attributes and their corresponding classes have been specified in Table \ref{tab:attrs}. The \textit{unsure} class for attributes gender and apparent age depict uncertainty on the annotator's end. For skin-tone annotation, the Fitzpatrick scale is used with class names \textit{very fair}, \textit{fair}, \textit{medium}, \textit{olive}, \textit{brown}, \textit{black}. Further, \textit{unsure} is used to capture skin type in grayscale images. The \textit{background} attribute is used to specify the background information for a given face. Classes include- \textit{plain/crowded}, \textit{in-focus/out-of-focus}. For attributes \textit{facial hair}, \textit{occlusion} and \textit{background}, multiple classes can hold true depending on the face image. The annotators are between 21-30 years of age with experience in face recognition research. 

\begin{table}[t]
\caption{\label{tab:attrs}Annotated attributes and their corresponding classes for the faces in the F2LA dataset used for study. The Multi-class column specifies whether a given face can belong to multiple classes for a given attribute.}
\centering
\small
\begin{tabular}{|l|l|c|}
\hline
\textbf{Attribute}                                            & \textbf{Classes}                                                                                        & \textbf{Multi-class} \\ \hline
Gender                                                        & male, female, unsure                                                                                    & \xmark                \\ \hline
Skin-tone                                                     & \begin{tabular}[c]{@{}l@{}}very fair, fair, medium, olive,\\ brown, black, unsure\end{tabular}          & \xmark                \\ \hline
Apparent Age                                                  & child, young, middle, old, unsure                                                                       & \xmark                \\ \hline
Blur                                                          & partial, heavy, no blur                                                                                 & \xmark                \\ \hline
Illumination                                                  & dim, bright, normal                                                                                     & \xmark                \\ \hline
\begin{tabular}[c]{@{}l@{}}Facial \\ Orientation\end{tabular} & frontal, semi-frontal, profile                                                                          & \xmark                \\ \hline
Facial Hair                                                   & \begin{tabular}[c]{@{}l@{}}mustache, goatee, beard, \\ no facial hair\end{tabular}                      & \checkmark            \\ \hline
Occlusion                                                     & \begin{tabular}[c]{@{}l@{}}eyes, forehead, mouth, nose,  \\ chin, cheek(s), facial boundary\end{tabular} & \checkmark            \\ \hline
Background                                                    & \begin{tabular}[c]{@{}l@{}}plain, crowded, in-focus, \\ out-of-focus\end{tabular}                       & \checkmark            \\ \hline
\begin{tabular}[c]{@{}l@{}}Color \\ Properties\end{tabular}   & grayscale, RGB                                                                                          & \xmark                \\ \hline
\end{tabular}
\end{table}

\begin{figure}[t]       
  \centering {\includegraphics[width=0.45\textwidth]{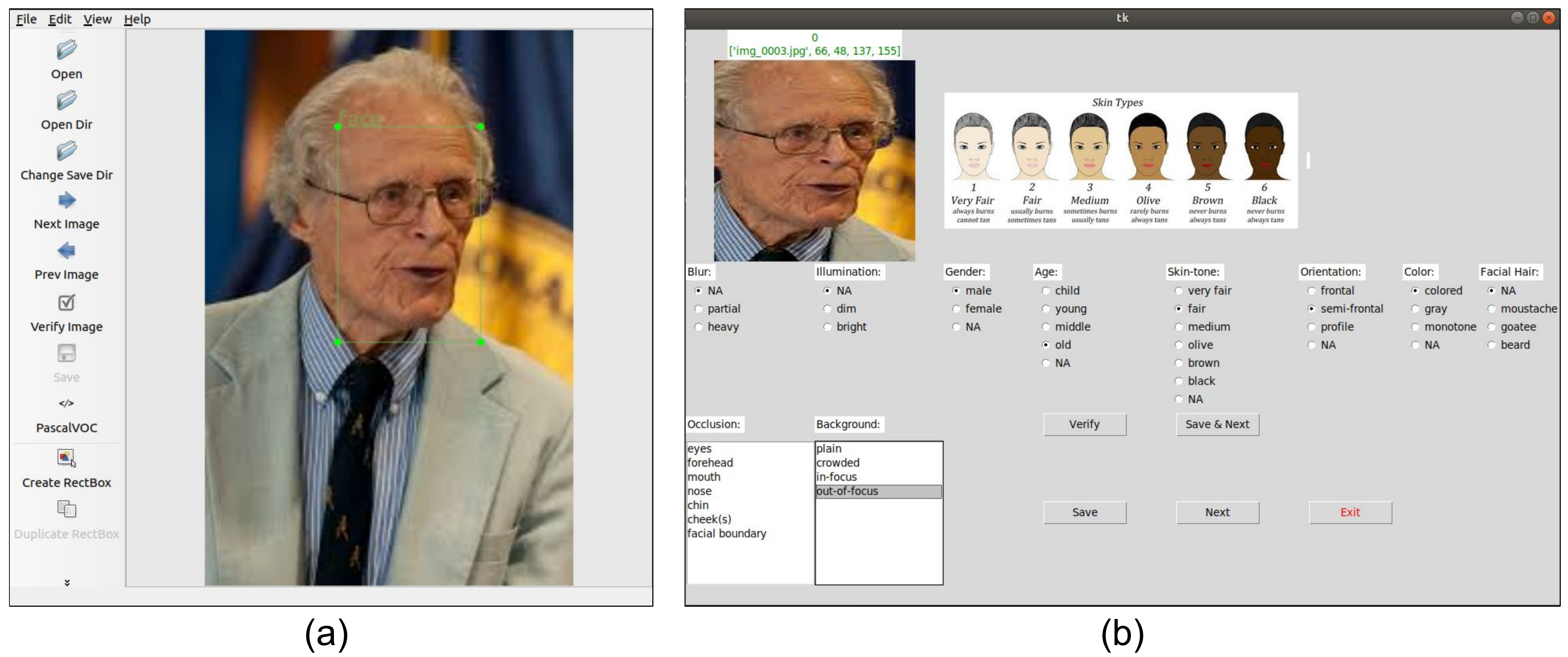}} 
    \caption{Tools used for (a) marking face regions with LabelImg, and (b) annotating attributes using created UI.}
    \label{anno_tools}
    \vspace{-10pt}
\end{figure}

\noindent \textbf{Data Annotation:} Each image is annotated in a two-step fashion as shown in Fig. \ref{anno_tools}. In the first step, all the faces from an image are cropped and annotated for a set of attributes described in Table \ref{tab:attrs}. The first step in the annotation process is face region localization. We used the freely available annotation tool, LabelImg \cite{labelimg} for this purpose. For the next step, we designed a user interface (UI) to facilitate the annotation of images.

\begin{figure}[t]
\centering
\includegraphics[width = 0.5\textwidth]{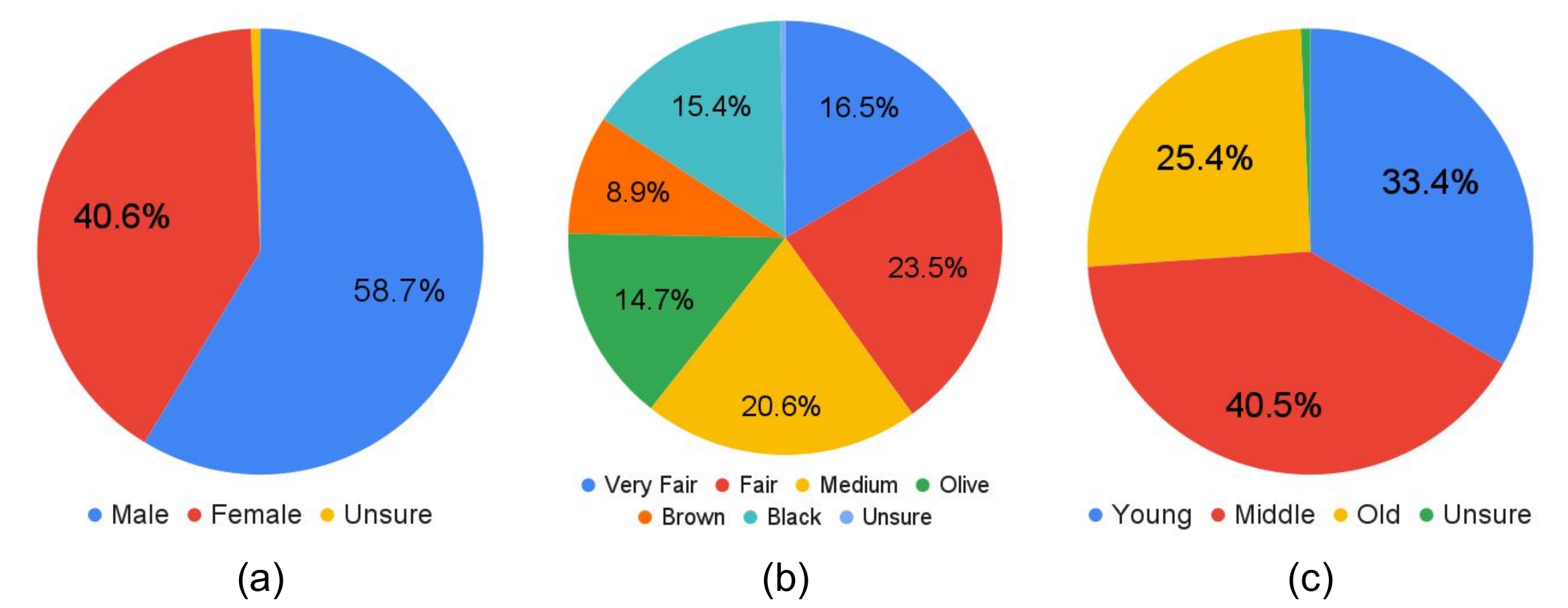}
\caption{Distribution of faces across (a) gender, (b) skin-tone, and (c) apparent age attributes on the test set.}
\label{fig:ImageDistribution}
\vspace{-10pt}
\end{figure}

\noindent \textbf{Dataset Protocol:}
\noindent The dataset consists of a total of 1200 images encompassing 1774 faces. The dataset is divided into train and test sets containing 1000 and 200 images with 1486 and 288 faces, respectively. Since the dataset contains ten annotations per face, we attempt to ensure as little skew as possible across the sensitive attributes (gender, skin-tone, age) in the test set. The distribution of faces across these attributes in the test set are shown in Fig. \ref{fig:ImageDistribution}. 

\subsection{Experimental Protocol}
\noindent We evaluate and estimate the performance and fairness of existing face detectors through two experiments. In the \textbf{first experiment}, the pre-trained models of existing face detectors are utilized. Since face detection models are often used as a part of pre-processing in many applications, the evaluation of pre-trained models aids in the estimation of possibly overlooked bias. This experiment is performed with four pre-trained face detectors, namely, MTCNN~\cite{mtcnn}, BlazeFace~\cite{blazeface}, DSFD~\cite{dsfd}, and RetinaFace~\cite{retinaface}. To further explore the prevalence of bias, we perform the \textbf{second experiment} in which we apply a transfer learning approach.  This experiment is performed to estimate performance improvement of existing face detectors on the proposed dataset. For this experiment, we employ the popular MTCNN model~\cite{mtcnn}, selected due to its lightweight architecture, suitable for finetuning with 1000 images.

\subsection{Details of Pre-trained Models}
\noindent Four pre-trained face detectors have been used for face detection. While MTCNN uses separate region-proposal and face localization networks, BlazeFace and RetinaFace employ the Single Shot Detector (SSD) design of training one-stage face detection frameworks end-to-end~\cite{liu2016ssd}.

\noindent \textbf{MTCNN~\cite{mtcnn}: } The MTCNN model is one of the earliest and most popular face detectors. It consists of three light-weight models namely P-Net, R-Net and O-Net used for region proposal refinement, bounding box localization and landmark localization, respectively. 

\noindent \textbf{BlazeFace~\cite{blazeface}: } The BlazeFace model is designed by Google to be a lightweight model suitable for GPU inference in mobiles. Its wide applicability makes it an essential candidate for evaluation. The pre-trained model used in this work is made available by Google's MediaPipe. 

\noindent \textbf{RetinaFace~\cite{retinaface}}: The RetinaFace detector used extra supervision through facial landmarks and employed self-supervised learning in addition to traditional bounding box regression. It is among the most widely implemented single-stage face detection  technique.

\noindent \textbf{DSFD~\cite{dsfd}: } The DSFD detector utilizes a Feature Enhance Module in addition to the single-shot detection of SSD models. It is among the best performing face detection models on several benchmark databases.

\begin{figure*}[t]
\centering
\includegraphics[scale = 0.55]{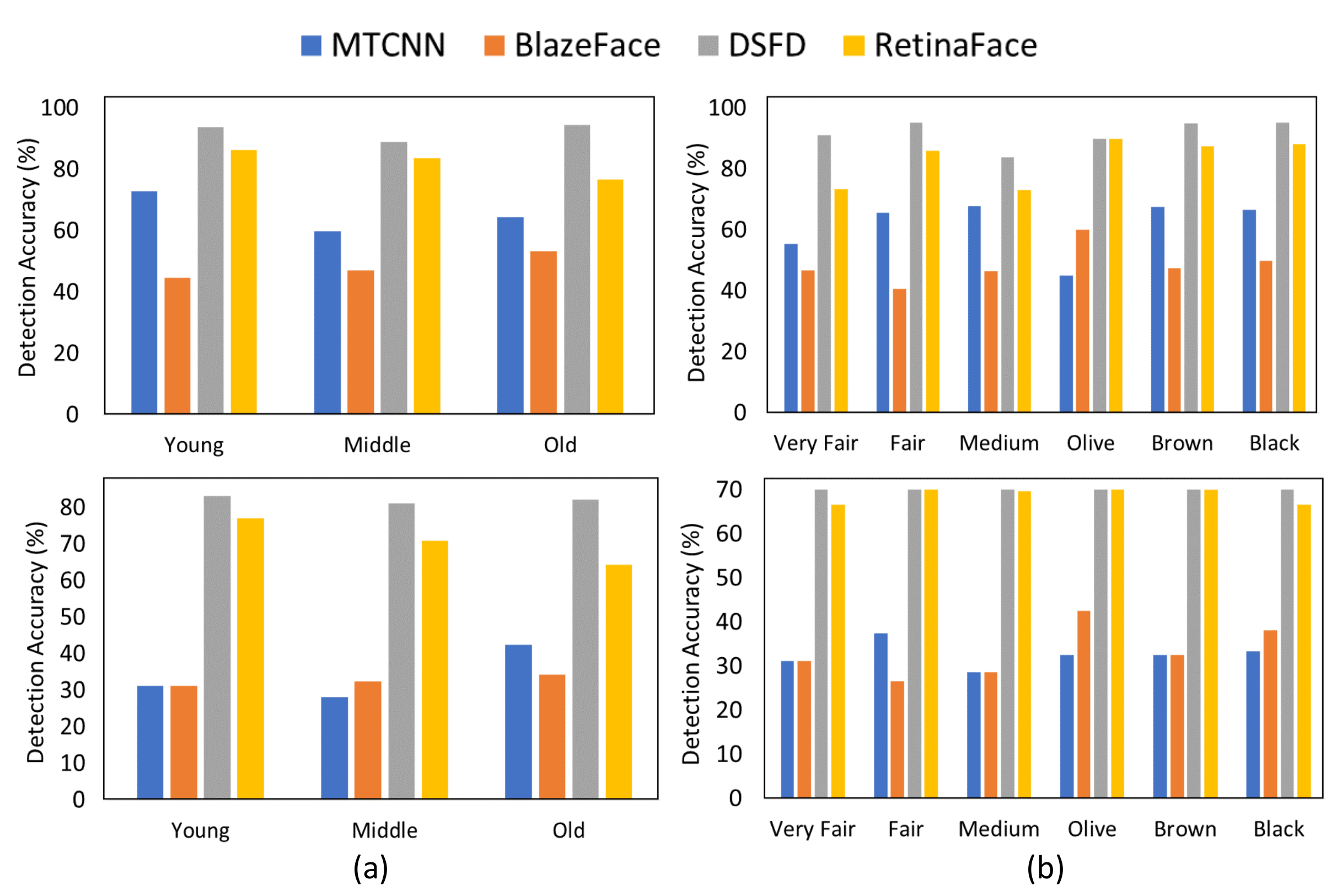}
\caption{Bar graph summarizing the performance of models across (a) age and (b) skin tone. The top row corresponds to $t$=0.6 and bottom row to $t$=0.7.}
\label{fig:ResGenSkin}
\vspace{-10pt}
\end{figure*}

\begin{table}[t]
\centering
\scriptsize
\renewcommand{\arraystretch}{1.2}
\caption{Detection accuracy (\%) across gender subgroups at different IoU thresholds $t$ using pre-trained models.}
\label{tab:gender_baseline}
\begin{tabular}{|c|l|c|c|c|c|}
\hline
$t$                    & \textbf{Gender}    & \textbf{MTCNN} & \textbf{BlazeFace} & \textbf{DSFD}  & \textbf{RetinaFace} \\ \hline
\multirow{4}{*}{\rotatebox{90}{0.5}} & Female             & 85.47          & 70.94              & 99.15          & 90.60               \\
                     & Male               & 72.78          & 62.13              & 93.49          & 85.80               \\ \cline{2-6} 
                     & \textit{Avg. Acc.} & \textit{79.13} & \textit{66.54}     & \textit{96.32} & \textit{88.20}      \\ \cline{2-6} 
                     & \textit{Disparity} & \textit{6.34}  & \textit{4.40}      & \textit{2.83}  & \textit{2.40}       \\ \hline
\multirow{4}{*}{\rotatebox{90}{0.6}} & Female             & 70.94          & 52.99              & 95.73          & 86.32               \\
                     & Male               & 56.21          & 44.38              & 88.76          & 79.88               \\ \cline{2-6} 
                     & \textit{Avg. Acc.} & \textit{63.58} & \textit{48.69}     & \textit{92.25} & \textit{83.10}      \\ \cline{2-6} 
                     & \textit{Disparity} & \textit{7.37}  & \textit{4.31}      & \textit{3.49}  & \textit{3.22}       \\ \hline
\end{tabular}
\vspace{-7pt}
\end{table}

\subsection{Evaluation Metrics}
\noindent For evaluating the performance of the detectors, we calculate the \textit{detection accuracy}. The detection accuracy for a set is computed as the ratio of correctly detected faces and the total number of faces in the set. A face is considered to be detected correctly if the predicted bounding box localization overlaps with the ground-truth annotation. This overlap is calculated using the \textit{IoU} (Intersection over Union) metric. The results are evaluated at multiple IoU thresholds $t$= 0.5, 0.6, and 0.7. To estimate the \textit{disparity} in performance across subgroups, we calculate standard deviation of the detection accuracies across the different classes. A high performance gap of the model across different classes will result in a high disparity value for that attribute, indicating higher bias in the model prediction.\\

\noindent \textit{Implementation Details:} For the pre-trained models, we use an MTCNN model pre-trained on the WIDERFACE dataset. A PyTorch/MXnet implementation of the paper is used. For BlazeFace, a PyTorch version of the model is used. The pre-trained weights for the frontal camera model are used in this paper for evaluation. For inferencing the DSFD detector, we employ an optimized version of the detector implemented in PyTorch. This model is pre-trained on the WIDERFACE dataset. Lastly, the RetinaFace detector used in this work is built with a ResNet50 backbone and also pre-trained on the WIDERFACE dataset. For the fine-tuning experiments, the MTCNN model is used. MTCNN contains the region-proposal refinement network P-Net and the bounding box localization network R-Net. We use the pre-trained P-Net and fine-tune the R-Net. Besides fine-tuning on the entire training set, we also fine-tune the network on three subsets of the train set which are balanced across the gender, skin-tone and apparent age attributes, respectively. All parameters except those of the last fully-connected layer of the R-Net are frozen and the model is trained for 20 epochs on the train set. The Adam optimizer is used with an initial learning rate of 0.001. All the experiments are performed on a workstation with Intel Xeon processor, having 128 GB RAM and an NVIDIA RTX-3090 GPU with 24 GB memory.


\section{Experimental Results and Analysis}

\noindent To estimate the bias present in different face detection models, we study their performance on gender, skin-tone and apparent age attributes present in the F2LA dataset. We further analyze some of the factors impacting the performance of the pre-trained detectors.

\begin{table}[t]
\scriptsize
\caption{Detection accuracies(\%) and disparity observed with the MTCNN across gender subgroups at IoU threshold $t$=0.6. Columns \textit{Complete} and \textit{Balanced} denote performance after fine-tuning the model with complete and gender-balanced training (sub)set, respectively. }
\renewcommand{\arraystretch}{1.2}
\centering
\label{tab:gender_mtcnn}
\begin{tabular}{|c|c|c|c|}
\hline
\textbf{Gender}    & \textbf{Pre-trained} & \textbf{Complete} & \textbf{Balanced} \\ \hline
Female             & 70.94                & 94.87             & 94.87             \\
Male               & 56.21                & 77.51             & 78.11             \\ \hline
\textit{Avg. Acc.} & \textit{63.58}       & \textit{86.19}    & \textit{86.49}    \\ \hline
\textit{Disparity} & \textit{7.36}        & \textit{8.68}     & \textit{8.38}     \\ \hline
\end{tabular}
\end{table}

\begin{figure*}[t]
\centering
\includegraphics[width=\textwidth]{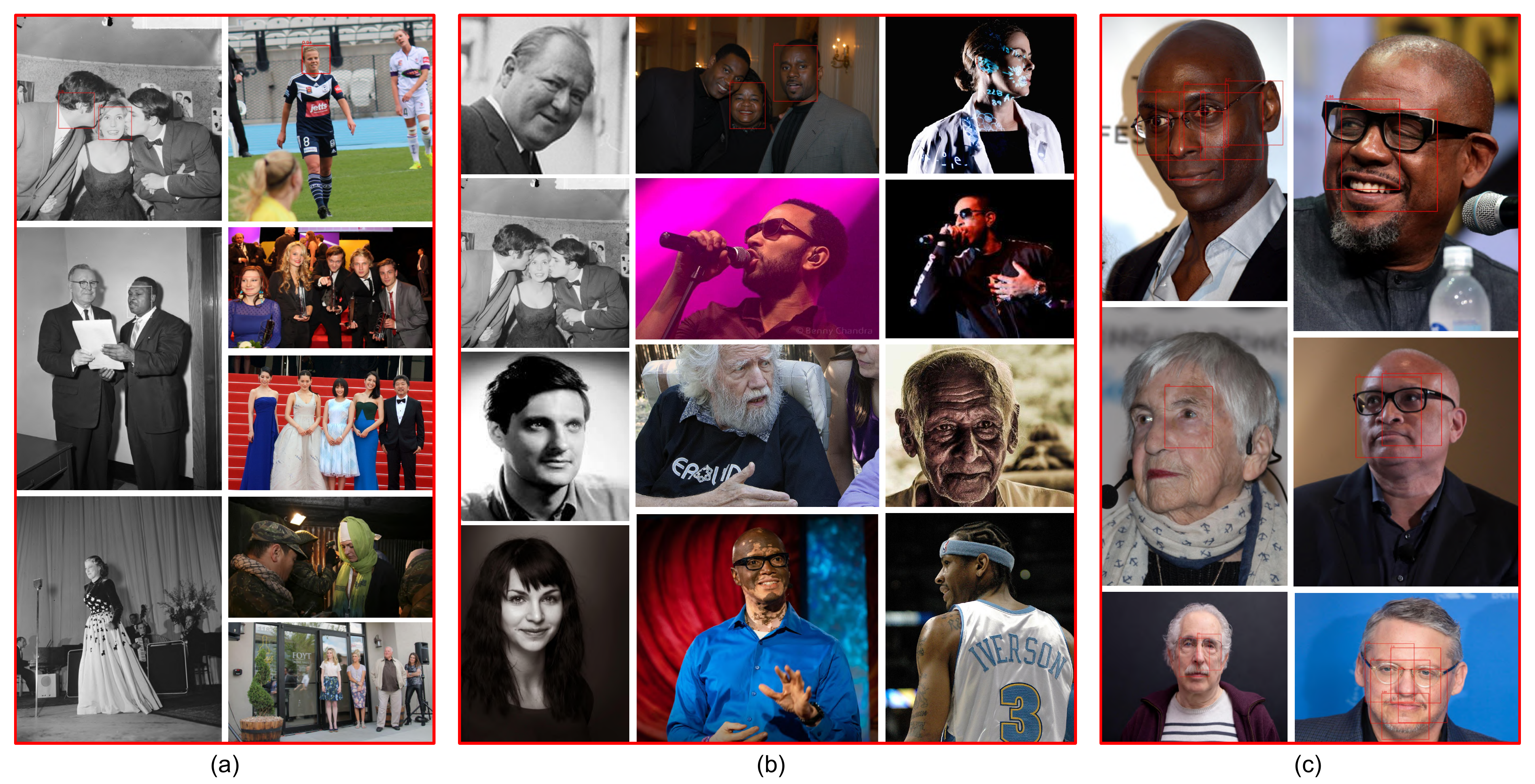}
\caption{Some of the samples missed by (a) BlazeFace, (b) MTCNN, and (c) RetinaFace face detectors.}
\label{fig:analysis}
\vspace{-10pt}
\end{figure*}

\subsection{Performance of Deep Models}
\noindent The pre-trained face detection models are evaluated across different demographic subgroups on the F2LA dataset. For \textit{gender subgroups}, the performance of the four pre-trained models has been provided in Table \ref{tab:gender_baseline}. At IoU threshold $t$=0.5, a wide gap exists between the detection accuracy for Male and Female. We observe a disparity of nearly 6.34\% and 2.40\% across gender subgroups in MTCNN and RetinaFace, respectively. Similar observations can be made for disparity at $t$=0.7. At $t$=0.6, a disparity of 8.44\%, 5.85\%, and 6.99\% is observed across different skin tones for MTCNN, Blazeface and RetinaFace detectors, respectively. This can be observed from Fig. \ref{fig:ResGenSkin} which showcases the difference in performance of pre-trained models across different age and skin-tone subgroups. \textit{Throughout all the experiments, a disparity in performance across the subgroups is observed consistently. While different models favor/disfavor different subgroups, the detection accuracy for the \textbf{male} subgroup is consistently low across different models.}

\begin{table}[t]
\caption{Disparities(\%) observed in detection accuracies at different IoU thresholds \textit{t} after fine-tuning the MTCNN model.}
\centering
\scriptsize
\renewcommand{\arraystretch}{1.2}
\label{tab:disparity_finetuned}
\begin{tabular}{|c|ccc|ccc|}
\hline
$t$                  & \multicolumn{3}{c|}{\textbf{0.5}}                                   & \multicolumn{3}{c|}{\textbf{0.6}}                                   \\ \hline
Attribute          & \multicolumn{1}{c|}{Age}  & \multicolumn{1}{c|}{Skin-tone} & Gender & \multicolumn{1}{c|}{Age}  & \multicolumn{1}{c|}{Skin-tone} & Gender \\ \hline
\textit{Disparity} & \multicolumn{1}{c|}{5.10} & \multicolumn{1}{c|}{4.36}      & 5.85   & \multicolumn{1}{c|}{3.72} & \multicolumn{1}{c|}{6.54}      & 8.68   \\ \hline
\end{tabular}
\end{table}

\noindent A major boost in performance ($>$20\%) is observed after fine-tuning the model with F2LA's complete training set as shown in Table \ref{tab:gender_mtcnn}. However, the disparity in performance across gender persists. Further, from Table \ref{tab:disparity_finetuned}, we observe that these disparities exist for other demographic subgroups such as skin-tone and age as well. On fine-tuning using an age-balanced split, we observe that the disparity reduces from \textit{6.13\%} to \textit{3.16\%} at $t$=0.7. Similarly, across the skin-tone subgroups, we observe a significant reduction in disparity from \textit{8.44\%} to \textit{5.60\%} with a skin-balanced fine-tuning of the network. \textit{While balanced training reduces the disparity in certain cases, a large gap in performance across subgroups still persists.}


\begin{table}[t]
\caption{Detection accuracies (\%) and their disparity observed across confounding factors such as Facial Orientation and Illumination at IoU threshold $t$=0.6. }
\centering
\scriptsize
\renewcommand{\arraystretch}{1.2}
\label{tab:otherfactors}
\begin{tabular}{|l|c|c|c|c|}
\hline
\textbf{Orientation}  & \textbf{MTCNN} & \textbf{BlazeFace} & \textbf{DSFD} & \textbf{RetinaFace} \\ \hline
Frontal               & 65.85          & 53.66              & 92.68         & 81.71               \\ \hline
Profile               & 36.00          & 20.00              & 80.00         & 76.00               \\ \hline
Semi-frontal          & 61.62          & 44.44              & 92.93         & 85.86               \\ \hline
\textit{Disparity}    & \textit{13.19} & \textit{14.20}     & \textit{6.04} & \textit{4.04}       \\ \hline \hline
\textbf{Illumination} & \textbf{MTCNN} & \textbf{BlazeFace} & \textbf{DSFD} & \textbf{RetinaFace} \\ \hline
Bright                & 57.14          & 50.00              & 92.86         & 92.86               \\ \hline
DIm                   & 44.00          & 32.00              & 92.00         & 92.00               \\ \hline
Normal                & 63.86          & 49.00              & 91.57         & 81.12               \\ \hline
\textit{Disparity}    & \textit{8.25}  & \textit{8.26}      & \textit{0.54} & \textit{5.34}       \\ \hline
\end{tabular}
\end{table}




\begin{figure*}[t]
\centering
\includegraphics[width=0.8\textwidth]{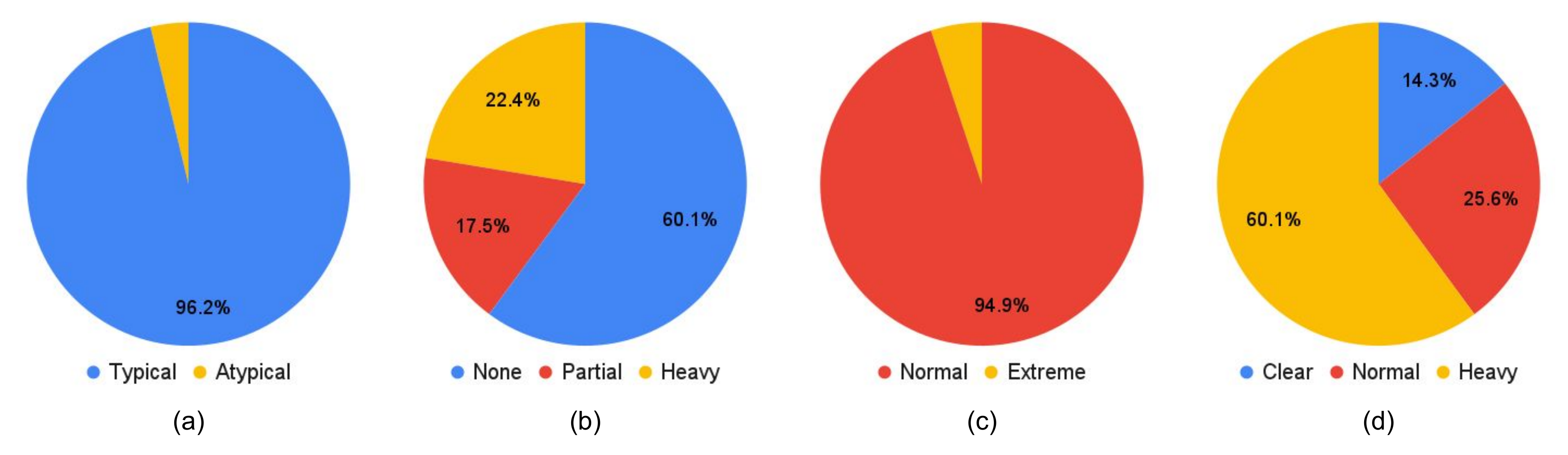}
\caption{Distribution in the training set of WIDERFACE dataset across the attributes- (a) pose, (b) occlusion, (c) illumination, and (d) blur.}
\label{fig:widerfdist}
\end{figure*}

\begin{figure*}[!h]
\centering
\includegraphics[scale = 0.85]{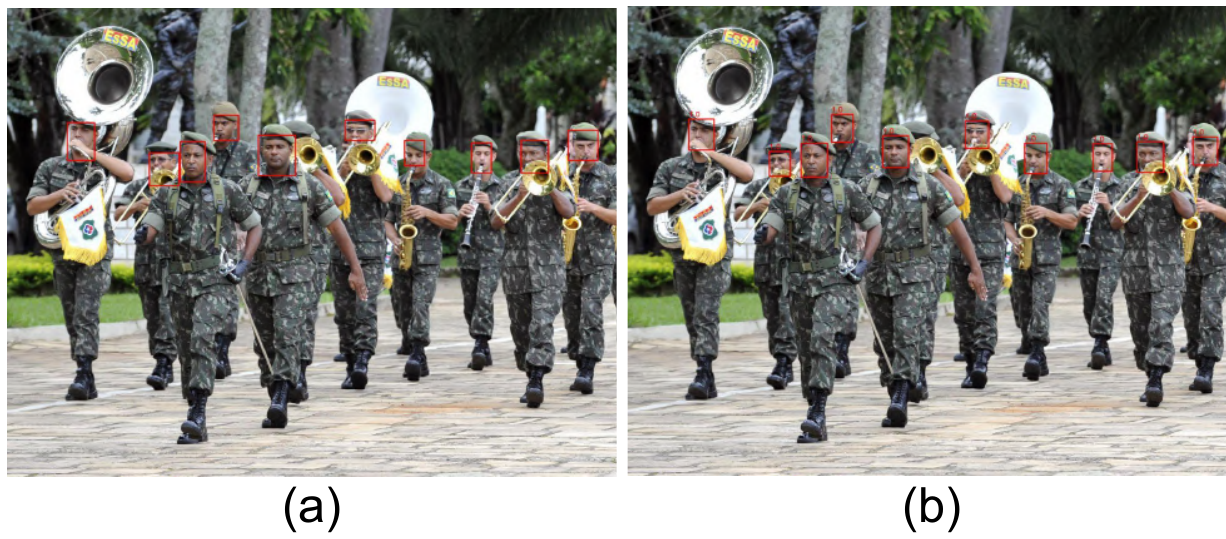}
\caption{Bounding boxes (a) in ground-truth, and (b) predicted by RetinaFace.}
\label{fig:iou}
\end{figure*}


\subsection{Analysis}
\noindent In this section, we study the impact of different factors on the performance of the face detection models. Fig. \ref{fig:analysis} presents some of the samples missed by the BlazeFace, MTCNN and RetinaFace models. \\

\noindent \textbf{Role of Demography:} We observe that some detection models perform poorly over certain demographic subgroups (Refer Table \ref{tab:gender_baseline} and Figure \ref{fig:ResGenSkin}). For example, the MTCNN model performs poorly on individuals belonging to \textit{old} and \textit{male} subgroup (Fig. \ref{fig:analysis}(b)). Similar observations have been made for the BlazeFace model (Fig. \ref{fig:analysis}(a)).\\

\noindent \textbf{Role of Other Confounding Factors: }
While analyzing results from the pre-trained models, we observe that the MTCNN and BlazeFace models fail to detect faces on grayscale images. Some samples are shown in Fig. \ref{fig:analysis}(a) and (b). Further, the BlazeFace model fails to detect small faces in an image. Conversely, the RetinaFace model fails to generalize on large faces (Fig. \ref{fig:analysis}(c)). Further, factors such as illumination and occlusion may play an important role in face detection. We observe disparate performance of the pre-trained models across the \textit{facial orientation} and \textit{illumination} attributes in Table \ref{tab:otherfactors}. Faces present with high illumination and frontal orientation are easily detected compared to those which are dimly illuminated and have profile view. \\


\noindent \textbf{Distribution of the WIDERFACE dataset: }
The bias in performance of models is impacted by the data it is trained on. Since the MTCNN, DSFD, and RetinaFace are trained on the same dataset- the WIDER FACE dataset~\cite{widerface}, studying the distribution of the dataset offers insight into the performance of the models. The WIDER FACE dataset contains 32,203 images with 393,703 faces with a 40\%/10\%/50\% split for training, validation, and testing. It is a widely popular large-scale dataset containing images under 62 different event categories taken from the WIDER dataset~\cite{xiong2015recognize}. Each face contains annotations for the following attributes- pose (typical, atypical), occlusion (none, partial, heavy), illumination (normal, extreme), and blur (clear, normal blur, heavy blur). The distribution of the images in the training set are shown in Fig. \ref{fig:widerfdist}. It is evident from the distribution that the train dataset is skewed towards \textit{Typical} poses and \textit{Normal} illumination. The results in Table \ref{tab:otherfactors} showcase higher performance of models for faces with these attributes in the F2LA dataset\footnote{The \textit{Typical} pose and \textit{Normal} illumination attributes correspond to the \textit{Frontal} facial orientation and \textit{Normal} illumination attributes in F2LA.}. Therefore, we can safely deduce that \textit{a skew in training data translates into model performance}. Further, the interplay between such confounding factors may play a role in the overall fairness of models. \\

\noindent \textbf{Validity of the IoU metric: }
On manual inspection of the faces detected by the DSFD and RetinaFace models, it is discovered that these models provided good detection results. \textit{So, why do we observe a disparity in Table \ref{tab:gender_baseline}?} We observe that while ground-truth annotations are done loosely, the predicted bounding boxes are tight (Fig. \ref{fig:iou}). This leads to a reduction in overlap leading to low performance at threshold $t$=0.5. The threshold $t$ selected for determining ideal overlap can lead to false negatives especially in cases where the area of interest is small, and has been shown to destabilize the performance~\cite{reinke2021common}. The limitations of the box IoU metric have been discussed in the literature where a lack of spatial information (compared to a mask IoU) leads to incorrect predictions~\cite{reinke2021common}. Since the image presented in Fig. \ref{fig:iou} contains faces of the \textit{male} subgroup and the bounding box area is small, the overall performance for \textit{males} in the test set is severely affected in this context.


\section{Discussion and Future Work}
\noindent In this research, we attempt to understand when and why face detection algorithms fail. By studying the impact of different factors on existing face detection algorithms, we also strive to ascertain whether face detection models are biased. The observations are as following:
\begin{itemize}
    \item While \textit{we experimentally detect disparate performance}, we observe that the interplay of various factors plays an important role. It is imperative to \textit{analyze the performance of models in a holistic manner} and not draw conclusions solely based on demographic annotation.
    \item Role of confounding factors such as scale of face, color properties, illumination and pose may be misattributed to gender or other kinds of bias. While evaluating bias, \textit{role of non-demographic factors should be evaluated}.
    \item We also note the limitations of the Intersection over Union metric for assessing face detection performance, and \textit{advise careful perusal of loosely versus tightly cropped faces} in the data. Several variations of IoU focusing on efficient training have been proposed in recent years~\cite{rezatofighi2019generalized, zheng2020distance}, however, a greater emphasis on IoU as an evaluation metric is also required.
\end{itemize}
\noindent  In the future, more effort is required to study face detection bias in a localization setting. We believe that the web-curated F2LA dataset collected across ten attributes for this study can help to broaden the scope for exploring issues in face detection pertaining to various confounding factors, including demography. 

\section{Acknowledgements}
S. Mittal is partially supported by the UGC-Net JRF Fellowship and IBM Fellowship. K. Thakral is partly supported by the PMRF Fellowship. M. Vatsa is partially supported through the Swarnajayanti Fellowship.

{\small
\bibliographystyle{ieee}
\bibliography{refs}

\begin{thebibliography}{10}\itemsep=-1pt

\bibitem{labelimg}
Tlabelimg. git code (2015).
\newblock \url{https://github.com/tzutalin/labelImg}, 2021.
\newblock Online; accessed 17 September 2021.

\bibitem{amazon}
ACLU.
\newblock Amazon’s face recognition falsely matched 28 members of congress
  with mugshots.
\newblock \url{https://tinyurl.com/2p87j4wu}, 2018.
\newblock Online; accessed 17 Feb 2022.

\bibitem{amini2019uncovering}
A.~Amini, A.~P. Soleimany, W.~Schwarting, S.~N. Bhatia, and D.~Rus.
\newblock Uncovering and mitigating algorithmic bias through learned latent
  structure.
\newblock In {\em AAAI/ACM Conference on AI, Ethics, and Society}, pages
  289--295, 2019.

\bibitem{blazeface}
V.~Bazarevsky, Y.~Kartynnik, A.~Vakunov, K.~Raveendran, and M.~Grundmann.
\newblock Blazeface: Sub-millisecond neural face detection on mobile gpus.
\newblock {\em arXiv preprint arXiv:1907.05047}, 2019.

\bibitem{gendershades}
J.~Buolamwini and T.~Gebru.
\newblock Gender shades: Intersectional accuracy disparities in commercial
  gender classification.
\newblock In {\em Conference on fairness, accountability and transparency},
  pages 77--91. PMLR, 2018.

\bibitem{retinaface}
J.~Deng, J.~Guo, E.~Ververas, I.~Kotsia, and S.~Zafeiriou.
\newblock Retinaface: Single-shot multi-level face localisation in the wild.
\newblock In {\em IEEE/CVF conference on Computer Vision and Pattern
  Recognition}, pages 5203--5212, 2020.

\bibitem{arcface}
J.~Deng, J.~Guo, J.~Yang, N.~Xue, I.~Kotsia, and S.~Zafeiriou.
\newblock Arcface: Additive angular margin loss for deep face recognition.
\newblock {\em IEEE Transactions on Pattern Analysis and Machine Intelligence},
  44(10):5962--5979, 2022.

\bibitem{drozdowski2020demographic}
P.~Drozdowski, C.~Rathgeb, A.~Dantcheva, N.~Damer, and C.~Busch.
\newblock Demographic bias in biometrics: A survey on an emerging challenge.
\newblock {\em IEEE Transactions on Technology and Society}, 1(2):89--103,
  2020.

\bibitem{fddb}
V.~Jain and E.~Learned-Miller.
\newblock Fddb: A benchmark for face detection in unconstrained settings.
\newblock Technical report, UMass Amherst technical report, 2010.

\bibitem{karkkainen2021fairface}
K.~Karkkainen and J.~Joo.
\newblock Fairface: Face attribute dataset for balanced race, gender, and age
  for bias measurement and mitigation.
\newblock In {\em IEEE/CVF Winter Conference on Applications of Computer
  Vision}, pages 1548--1558, 2021.

\bibitem{kortylewski2019analyzing}
A.~Kortylewski, B.~Egger, A.~Schneider, T.~Gerig, A.~Morel-Forster, and
  T.~Vetter.
\newblock Analyzing and reducing the damage of dataset bias to face recognition
  with synthetic data.
\newblock In {\em IEEE/CVF Conference on Computer Vision and Pattern
  Recognition Workshops}, pages 0--0, 2019.

\bibitem{dsfd}
J.~Li, Y.~Wang, C.~Wang, Y.~Tai, J.~Qian, J.~Yang, C.~Wang, J.~Li, and
  F.~Huang.
\newblock Dsfd: dual shot face detector.
\newblock In {\em IEEE/CVF Conference on Computer Vision and Pattern
  Recognition}, pages 5060--5069, 2019.

\bibitem{liu2016ssd}
W.~Liu, D.~Anguelov, D.~Erhan, C.~Szegedy, S.~Reed, C.-Y. Fu, and A.~C. Berg.
\newblock Ssd: Single shot multibox detector.
\newblock In {\em European Conference on Computer Vision}, pages 21--37.
  Springer, 2016.

\bibitem{lcsee}
A.~Majumdar, R.~Singh, and M.~Vatsa.
\newblock Face verification via class sparsity based supervised encoding.
\newblock {\em IEEE Transactions on Pattern Analysis and Machine Intelligence},
  39(6):1273--1280, 2017.

\bibitem{majumdar2021unravelling}
P.~Majumdar, S.~Mittal, R.~Singh, and M.~Vatsa.
\newblock Unravelling the effect of image distortions for biased prediction of
  pre-trained face recognition models.
\newblock In {\em IEEE/CVF International Conference on Computer Vision
  Workshops}, pages 3786--3795. IEEE, 2021.

\bibitem{majumdar2021attention}
P.~Majumdar, R.~Singh, and M.~Vatsa.
\newblock Attention aware debiasing for unbiased model prediction.
\newblock In {\em IEEE/CVF International Conference on Computer Vision}, pages
  4133--4141, 2021.

\bibitem{propublica}
ProPublica.
\newblock Machine bias.
\newblock
  \url{https://www.propublica.org/article/machine-bias-risk-assessments-in-criminal-sentencing},
  2016.
\newblock Online; accessed 17 Feb 2022.

\bibitem{reinke2021common}
A.~Reinke, M.~Eisenmann, M.~D. Tizabi, C.~H. Sudre, T.~R{\"a}dsch,
  M.~Antonelli, T.~Arbel, S.~Bakas, M.~J. Cardoso, V.~Cheplygina, et~al.
\newblock Common limitations of image processing metrics: A picture story.
\newblock {\em arXiv preprint arXiv:2104.05642}, 2021.

\bibitem{rezatofighi2019generalized}
H.~Rezatofighi, N.~Tsoi, J.~Gwak, A.~Sadeghian, I.~Reid, and S.~Savarese.
\newblock Generalized intersection over union: A metric and a loss for bounding
  box regression.
\newblock In {\em IEEE/CVF conference on computer vision and pattern
  recognition}, pages 658--666, 2019.

\bibitem{robinson2020face}
J.~P. Robinson, G.~Livitz, Y.~Henon, C.~Qin, Y.~Fu, and S.~Timoner.
\newblock Face recognition: too bias, or not too bias?
\newblock In {\em IEEE/CVF Conference on Computer Vision and Pattern
  Recognition Workshops}, pages 0--1, 2020.

\bibitem{aaai2020}
R.~Singh, A.~Agarwal, M.~Singh, S.~Nagpal, and M.~Vatsa.
\newblock On the robustness of face recognition algorithms against attacks and
  bias.
\newblock {\em AAAI Conference on Artificial Intelligence},
  34(09):13583--13589, 2020.

\bibitem{singh2022anatomizing}
R.~Singh, P.~Majumdar, S.~Mittal, and M.~Vatsa.
\newblock Anatomizing bias in facial analysis.
\newblock In {\em AAAI Conference on Artificial Intelligence}, volume~36, pages
  12351--12358, 2022.

\bibitem{Twitter}
J.~Vincent.
\newblock Twitter’s photo-cropping algorithm prefers young, beautiful, and
  light-skinned faces.
\newblock \url{https://tinyurl.com/xse4xmds}, 2021.
\newblock Online; accessed 27 August 2021.

\bibitem{wang2019racial}
M.~Wang, W.~Deng, J.~Hu, X.~Tao, and Y.~Huang.
\newblock Racial faces in the wild: Reducing racial bias by information
  maximization adaptation network.
\newblock In {\em IEEE/CVF International Conference on Computer Vision}, pages
  692--702, 2019.

\bibitem{xiong2015recognize}
Y.~Xiong, K.~Zhu, D.~Lin, and X.~Tang.
\newblock Recognize complex events from static images by fusing deep channels.
\newblock In {\em IEEE Conference on Computer Vision and Pattern Recognition},
  pages 1600--1609, 2015.

\bibitem{pascalface}
J.~Yan, X.~Zhang, Z.~Lei, and S.~Z. Li.
\newblock Face detection by structural models.
\newblock {\em Image and Vision Computing}, 32(10):790--799, 2014.

\bibitem{widerface}
S.~Yang, P.~Luo, C.-C. Loy, and X.~Tang.
\newblock Wider face: A face detection benchmark.
\newblock In {\em IEEE conference on computer vision and pattern recognition},
  pages 5525--5533, 2016.

\bibitem{mtcnn}
K.~Zhang, Z.~Zhang, Z.~Li, and Y.~Qiao.
\newblock Joint face detection and alignment using multitask cascaded
  convolutional networks.
\newblock {\em IEEE Signal Processing Letters}, 23(10):1499--1503, 2016.

\bibitem{zheng2020distance}
Z.~Zheng, P.~Wang, W.~Liu, J.~Li, R.~Ye, and D.~Ren.
\newblock Distance-iou loss: Faster and better learning for bounding box
  regression.
\newblock In {\em AAAI conference on artificial intelligence}, volume~34, pages
  12993--13000, 2020.

\bibitem{afw}
X.~Zhu and D.~Ramanan.
\newblock {Face detection, pose estimation, and landmark localization in the
  wild}.
\newblock In {\em IEEE Conference on Computer Vision and Pattern Recognition},
  pages 2879--2886. IEEE, 2012.

\end{thebibliography}
}

\end{document}